\documentclass{article}

\usepackage[nonatbib, final]{neurips_2021_ml4ad}

\usepackage[utf8]{inputenc} 
\usepackage[T1]{fontenc}    
\usepackage{hyperref}       
\usepackage{url}            
\usepackage{booktabs}       
\usepackage{amsfonts}       
\usepackage{nicefrac}       
\usepackage{microtype}      
\usepackage{xcolor}         
\usepackage{epsfig}
\usepackage{graphicx}
\usepackage{amsmath}
\usepackage{amssymb}
\usepackage{multirow}
\usepackage{wrapfig,lipsum,booktabs}
\usepackage[numbers]{natbib}

\newcommand{\figref}[1]{Fig.~\ref{#1}}

\newcommand{\equref}[1]{Equ.~(\ref{#1})}

\title{Object-Level Targeted Selection via Deep Template Matching}

\author{Suraj Kothawade \thanks{Work done while the author was at NVIDIA.} \\ University of Texas at Dallas \\ \texttt{suraj.kothawade@utdallas.edu}
\And
Donna Roy \\ NVIDIA \\ \texttt{donnar@nvidia.com}
\And
Michele Fenzi \\ NVIDIA \\ \texttt{mfenzi@nvidia.com}
\And
Elmar Haussmann \\ NVIDIA \\ \texttt{ehaussmann@nvidia.com}
\And
Jose M. Alvarez \\ NVIDIA \\ \texttt{josea@nvidia.com}
\And
Christoph Angerer \\ NVIDIA \\ \texttt{cangerer@nvidia.com}
}

\begin{document}

\maketitle

\begin{abstract}
  Retrieving images with objects that are semantically similar to objects of interest (OOI) in a query image has many practical use cases. A few examples include fixing failures like false negatives/positives of a learned model or mitigating class imbalance in a dataset. The targeted selection task requires finding the relevant data from a large-scale pool of unlabeled data. Manual mining at this scale is infeasible. Further, the OOI are often small and occupy less than 1\% of image area, are occluded, and co-exist with many semantically different objects in cluttered scenes. Existing semantic image retrieval methods often focus on mining for larger sized geographical landmarks, and/or require extra labeled data, such as images/image-pairs with similar objects, for mining images with generic objects. We propose a fast and robust template matching algorithm in the DNN feature space, that retrieves semantically similar images at the object-level from a large unlabeled pool of data. We project the region(s) around the OOI in the query image to the DNN feature space for use as the template. This enables our method to focus on the semantics of the OOI without requiring extra labeled data. In the context of  autonomous driving, we evaluate our system for targeted selection by using failure cases of object detectors as OOI. We demonstrate its efficacy on a large unlabeled dataset with 2.2M images and show high recall in mining for images with small-sized OOI. We compare our method against a well-known semantic image retrieval method, which also does not require extra labeled data. Lastly, we show that our method is flexible and retrieves images with one or more semantically different co-occurring OOI seamlessly.
\end{abstract}

\section{Introduction}
Retrieving images with objects that are semantically similar to objects of interest (OOI) in a query image, at a large-scale, is a fundamental task in computer vision. It has many practical applications, e.g. i) targeting consistent failure cases of trained models, ii) targeting class imbalance in object detection datasets where the imbalance can be due to various attributes like class, object size, etc. Such problems can be solved by adding relevant samples to training data. Our inspiring motivation was driven by targeted selection at an object-level, where the failure cases like false negatives/positives in production level object detectors for autonomous driving vehicles (AV) are used as OOI. These failure cases occur across different driving scenarios and conditions, and are often due to under-representation of such data in the training dataset. On the other hand, collecting unlabeled AV data at a large-scale is easily accomplished even with a small fleet of cars. But, manual mining for relevant data at this scale is not feasible. 

\begin{wrapfigure}{R}{0.50\textwidth}
    \centering
    \includegraphics[width=0.50\textwidth]{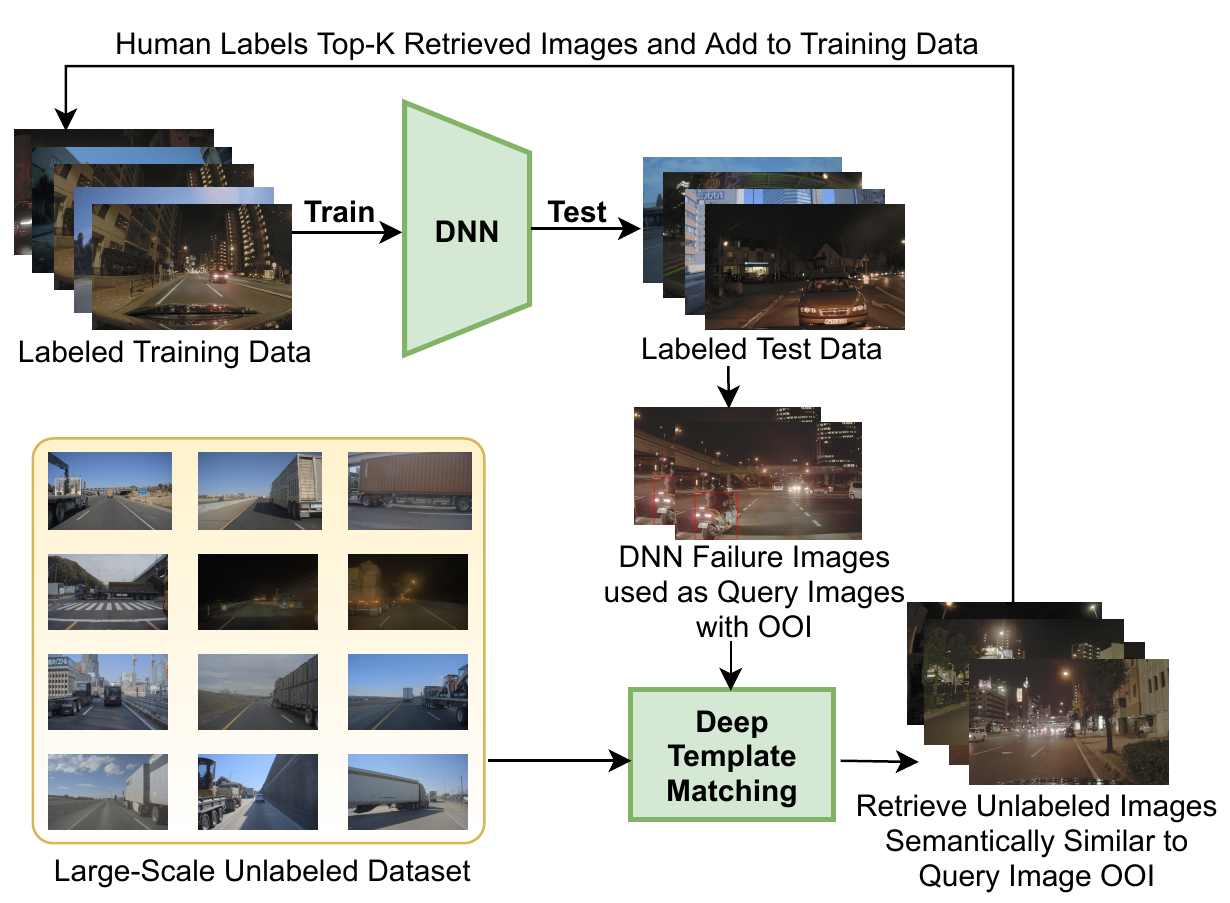}
    \caption{Motivating use case for object-level targeted selection via Deep Template Matching. DTM mines for semantically similar images to OOI containing failure cases like false negatives/positives. Top-k scored images can be labeled by a human and added to the training data.}
    \label{fig:mining_task}
    \vspace{-0.5cm}
\end{wrapfigure}

Thus, the task of object-level targeted selection involves automatically finding the relevant data with objects that are semantically similar to the OOI from a large-scale pool of unlabeled data, as shown in~\figref{fig:mining_task}. Further, these OOI are often small-sized and occupy less than 1\% of image area, are occluded, and co-exist with many semantically different objects in real-world cluttered scenes which makes the mining task challenging.

We formulate the object-level targeted selection task as a semantic \textit{sub-image} retrieval problem which takes as input a small set of query images (e.g., a few tens of query images) with bounding box annotations around one or more OOI in the scene and finds images from a large pool of unlabeled data with semantically similar objects. These annotations form the region of interest (ROI) in the query images. Typically, examples of these ROIs are objects which the object detector systematically failed to detect, e.g., motorcycles at night, bicycles mounted on cars, etc. Since the ROI is used as a template for search, and we use deep features to represent the template and image search space, we name our approach Deep Template Matching (DTM).

In the last decade, significant advances have been made in large-scale semantic image retrieval systems - from handcrafted features and indexing algorithms \cite{jegou2008hamming, lowe2004distinctive, philbin2007object, sivic2003video} to more recent methods based on DNNs for global descriptor learning \cite{arandjelovic2016netvlad, babenko2015aggregating, babenko2014neural, gong2014multi, gordo2017end, kalantidis2016cross, razavian2016visual, sharif2014cnn, teichmann2019detect, tolias2020learning, tolias2015particular}. 

The recent advances in DNN-based global descriptors for image retrieval tend to focus on mining for larger-sized geographical landmarks. Their performance is hindered in some challenging conditions observed in real-world AV datasets, such as small-sized objects of interest, occlusion, and heavy scene clutter. 

Various methods have been proposed in the literature to overcome this limitation, such as instance-level image retrieval methods in \cite{teichmann2019detect, gordo2017end, gordo2017beyond, li2018deep, tolias2020learning, yu2020fine} which tend to be computationally expensive, mining DNNs with attention modules in \cite{noh2017large, cao2020unifying, kim2018regional, ng2020solar} and deep metric learning \cite{akata2016multi, berman2019multigrain, cao2020enhancing, chen2019hybrid, wang2019multi, zhai2018classification} and neural graph \cite{juan2019graph} based methods which require extra labeled data to explicitly train the mining DNNs. Another line of related work is template matching techniques in \cite{ cheng2019qatm, dekel2015best, fuh1991motion, kat2018matching, korman2013fast, oron2017best, talmi2017template, tsai2002rotation} which solve the problem of finding a template patch in a sample image. However, these methods typically use low-level image features and are usually susceptible to geometric transformations (like rotation, translation, scale, etc.), illumination changes, occlusions, and background clutter. We describe related work in more details in section \ref{sec:2}.

\textbf{Our main contributions:} In this paper, we propose Deep Template Matching (DTM) - a fast and robust template matching algorithm in the DNN feature space, that retrieves semantically similar images at the object-level from a large unlabeled pool of data. We solve this by projecting the region(s) around the objects of interest in the query image to the DNN feature space for use as the template, and then computing a linear-time one-shot similarity score in the deep feature space. This enables our method to focus on the semantics of the objects of interest, even when the object is small-sized, amid occlusion and heavy scene clutter, without requiring extra labeled data, and is computationally cheap. We demonstrate the effectiveness of our method on a labeled dataset with 465k images and a large unlabeled dataset with 2.2M images. We compare DTM against a well-known semantic image retrieval method \cite{babenko2015aggregating} which also does not require extra labeled data. Lastly, we illustrate the flexibility of DTM for multiple queries without increasing computational complexity. 

\textbf{DTM has the following characteristics:} 1) \textit{Accuracy}: High recall values for queries with one or more objects of interest (i.e., templates) of any size and multiple semantic categories in real-world scenes with occlusion and heavy clutter. 2) \textit{Computational efficiency}: Quick mining time on the GPU which is essential for fast experimental turnaround time in production-level pipelines. DTM allows precomputing the image embeddings in DNN feature space offline, to speed up the scoring and search later. The computational complexity of DTM does not depend on the number or semantic category of the objects of interest in the query image. 3) \textit{Multi-template search in one shot}: Mining using multiple objects of interest belonging to one or more semantic categories co-occurring in the sample image is done seamlessly in one shot for the same computational complexity as a single object of interest. 4) \textit{Ability to preserve object size semantic information}: DTM mines for semantically similar samples of the same size as the query image ROI by zeroing out irrelevant features and maintaining the dimensionality of the original feature map (see \figref{fig:labeled_exp}a). Mining small objects from unlabeled data is critical, since many object detection datasets lack labels for small objects.

Our experiments show that DTM can successfully find images with semantically similar objects even when the objects of interest in the query image are quite small in size (e.g., occupies less than 0.3\% of area in input 2MP image), and there is occlusion and heavy scene clutter with multiple kinds of other objects, e.g., busy streets with many cars, pedestrians, etc., in the query and unlabeled pool of images. As discussed in section \ref{sec:2}, existing methods in the literature to solve this are either computationally expensive or require additional labeled training data. DTM is computationally cheap, 
works seamlessly for multiple co-occurring objects of the same or different semantic category, and does not require additional labeled training data. 

\section{Related Work} \label{sec:2}
The following two problems are closely related to object-level targeted selection.\\
\noindent{\textbf{Large-scale semantic image retrieval systems.}} Early retrieval systems used handcrafted features and indexing algorithms \cite{jegou2008hamming, lowe2004distinctive, philbin2007object, sivic2003video}. More recently, methods based on DNNs for global descriptor learning \cite{arandjelovic2016netvlad, babenko2015aggregating, babenko2014neural, gong2014multi, gordo2017end, kalantidis2016cross, razavian2016visual, sharif2014cnn, teichmann2019detect, tolias2020learning, tolias2015particular} have performed quite well in the literature. Typically, they use DNNs pre-trained on ImageNet as deep feature extractors, and focus on designing image representations suited for image retrieval on top of these features. These approaches fail when (i) the object of interest in the query image is small, or (ii) the scenes are heavily cluttered with multiple objects. Since these are very typical in real world highway and urban driving scenarios, such approaches did not work well for our AV use case. We implement the approach described in \cite{babenko2015aggregating} as baseline and demonstrate better performance of DTM over \cite{babenko2015aggregating} on our AV data. 

Some recent DNN based local features have been proposed for patch-level matching \cite{han2015matchnet, yi2016lift, zagoruyko2015learning}. However, these techniques focus on low-level image features like texture, and geometric information. They do not detect semantically meaningful features at the object-level.

In another line of work, \cite{teichmann2019detect, gordo2017end, gordo2017beyond, li2018deep, tolias2020learning, yu2020fine} apply an instance-level image retrieval scheme to find object/landmark instances among cluttered scenes. These methods focus on finding multiple viable region proposals in a single image, extracting deep local feature descriptors for each proposed region, and then aggregating these region-based local descriptors into one global descriptor for the entire image. Searching for viable region proposals and computing deep feature descriptors for each proposed region within a single sample image is non-trivial and computationally expensive. 

Approaches in \cite{noh2017large, cao2020unifying, kim2018regional, ng2020solar}, solve the problem of finding object/landmark instances in occluded/cluttered scenes by explicitly training the image retrieval DNN with labeled data to learn an attention module. These methods require extra labeled data for training the mining DNN, ranging from image-level annotations in \cite{noh2017large, cao2020unifying} to bounding box level annotations in \cite{ng2020solar}. Recent advances in deep metric learning \cite{akata2016multi, berman2019multigrain, cao2020enhancing, chen2019hybrid, wang2019multi, zhai2018classification} and neural graphs \cite{juan2019graph} also focus on this problem, but they also require extra labeled data either in terms of pairs or triplets of similar and dissimilar scenes or to populate the neural graph. DTM does not need extra labeled data, which is advantageous, since labeling is an expensive human-in-the-loop operation.

\noindent{\textbf{Template Matching.}} Another line of related work is template matching, which solves the problem of finding a template patch in a candidate window in the sample image.

Classical template matching techniques \cite{chen2003fast, elboher2013asymmetric, hel2013matching, ouyang2011performance} use sum-of-squared-differences (SSD) or normalized cross correlation (NCC) as the similarity measure in the input image feature space, and are quite sensitive to any variation in illumination and noise. More robust measures such as M-estimators\cite{chen2003fast, sibiryakov2011fast} or hamming based distance \cite{shin2007fast, pele2008robust} have been studied. However, all of these methods are not robust to real-world scenes since they account for only a strict rigid geometric transformation (only translation) during the measure computation. 
Several works in the parametric family attempt to overcome this problem. The approach in \cite{korman2013fast} uses 2D affine transformation to account for geometric differences between the template and the sample image. \cite{tian2012globally} addresses non-rigid transformations by parametric estimation of the distortion. These methods are susceptible to noise, occlusion and clutter. Moreover, they use a parametric approach, which is not required by DTM.  


More recently, several robust non-parametric template matching approaches were suggested to mitigate noise, occlusions and degradation. \cite{dekel2015best, talmi2017template} use  Nearest-Neighbor (NN) matches between features of the template and a sample image. The Best Buddies-Similarity measure in \cite{dekel2015best} focuses on nearest-neighbor matches to remove bad matches due to background pixels. The Deformable Diversity Similarity (DDS) in \cite{talmi2017template} considers possible template deformations and uses the diversity of nearest neighbors feature matches. Computing NN is computationally expensive even with optimized libraries. Further, in such algorithms one cannot exploit matrix properties that can enable scoring multiple images in a \textit{single shot}, as is done in DTM. \cite{kat2018matching} uses co-occurrence statistics to quantify the similarity between the template and a candidate window in the sample image. The co-occurrence matrix stores the count of the number of times two features appear together in an image in a fixed-sized heuristically-chosen window. The approach in \cite{cheng2019qatm} defines a bidirectional softmax based likelihood function as the score between the template and the sample image. In contrast, DTM is computationally efficient. It uses a single shot unidirectional scoring function (see \equref{eq:sc1}) which lends itself well to batched matrix operations. DTM does not require any heuristics as in \cite{huttenlocher1993comparing, kat2018matching} since the similarity between the template and sample image is computed in a patchwise fashion by projecting the template in the DNN feature space.

\section{Deep Template Matching}
In this section, we present and discuss the architecture and scoring mechanism of DTM in detail. 

\subsection{Architecture}

\begin{wrapfigure}{R}{0.50\textwidth}
    \centering
    \includegraphics[width=0.50\textwidth]{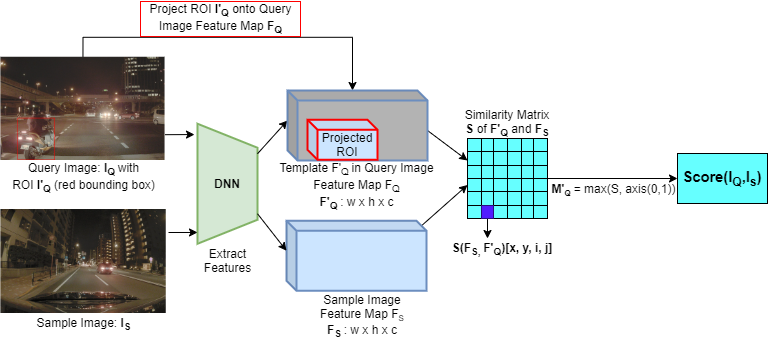}
    \caption{Architecture of DTM.}
    \label{fig:dtm_arch}
\end{wrapfigure}

Figure \ref{fig:dtm_arch} describes the end-to-end architecture of DTM. It takes as input a query image $I_Q$ with a bounding box around object(s) of interest (i.e., region of interest) and a set $A$ of unlabeled sample images. Let $I_Q'$ be the ROI in the query image. The goal is to find images $I_S$ from the set $A$ of unlabeled images which have regions semantically similar to the region of interest $I_Q'$. For simplicity, let $I_Q$ be a single query image with a single ROI $I_Q'$ and $I_S$ be a single sample image. We extract the features $F_Q$, $F_S \in \mathbb{R}^{w \times h \times c}$ for $I_Q$ and $I_S$ respectively, using a pre-trained object detector DNN, where the dimensions are denoted as, $w$: width, $h$: height and $c$: number of channels. Next, we project $I_Q'$ onto $F_Q$ to obtain $F'_Q \in \mathbb{R}^{w \times h \times c}$. We do so by linearly mapping the ROI $I_Q'$ onto $F_Q$ and zeroing out features that do not have a one-to-one correspondence with $I_Q'$ (see~\figref{fig:dtm_arch}).
Note that $F'_Q$ is the final representation of the template, which is used to compute similarity with all $F_S \in A$. Before using these features for computing similarity, we L2-normalize them along the channel dimension $c$. We discuss the benefits of doing this normalization in the next section. 

\subsection{Patchwise Similarity Computation}
\begin{wrapfigure}{R}{0.50\textwidth}
    \centering
    \includegraphics[width=0.50\textwidth]{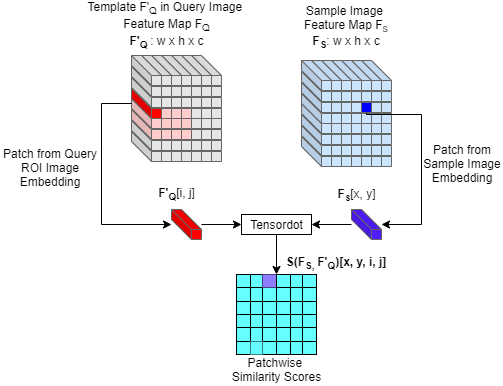}
    \caption{Patchwise Cosine Similarity.}
    \label{fig:dtm_score}
\end{wrapfigure}

In this section, we focus on computing the similarity matrix $S$ in~\figref{fig:dtm_score}. Consider feature tensors $F'_Q$ and $F_S$, where each spatial location is denoted by $F'_Q[i, j]$ and $F_S[x, y]$ respectively (see \figref{fig:dtm_score}). We define each of these tensors of depth $c$ as a \textit{patch} because they map to a certain region (patch) in the original image. Hence, both $F'_Q$ and $F_S$ have $w*h$ \textit{patchwise} feature vectors, each of length $c$.

Next, we compute patchwise cosine similarity scores between $F'_Q$ and $F_S$ by computing the similarity between $F'_Q[i, j]$ and $F_S[x, y]$ for all values of $x,i \in  [0,w-1]$ and $y, j \in [0, h-1]$. Note that $F'_Q[i, j], F_S[x,y] \in \mathbb{R}^{c}$. Strictly, the similarity score between the zeroed patches in $F'_Q$ and $F_S$ need not be computed, as it does not affect the final score in \equref{eq:sc1}. 

Intuitively, the goal is to find the score of the best matching sample patch feature for each query patch feature $F'_Q[i,j]$, at spatial location $(i, j)$. To compute this score, we use cosine similarity due to its computational efficiency. We require a score between every patch in the query image to every patch in the sample image, and compute the cosine similarity along the channel dimension. This results in a 4-D patchwise cosine similarity tensor $S \in \mathbb{R}^{w \times h \times w \times h}$ which is computed according to 
\begin{equation}\label{eq:fullsimcomp}
    S(F_S, F'_Q) = (F_S \cdot  F'_Q, axis=c)
\end{equation}

At a patch level, each element in $S$ at location $[x, y, i, j]$ is computed as 

\begin{equation}\label{eq:simcomp}
    S(F_S, F'_Q) [x, y, i, j] = \frac{F_S[x, y] \cdot  F'_Q[i, j]}{ |F_S[x, y]|_2  |F'_Q[i, j]|_2}
\end{equation}

\begin{wrapfigure}{R}{0.5\textwidth}
    \centering
    \includegraphics[width=0.5\textwidth]{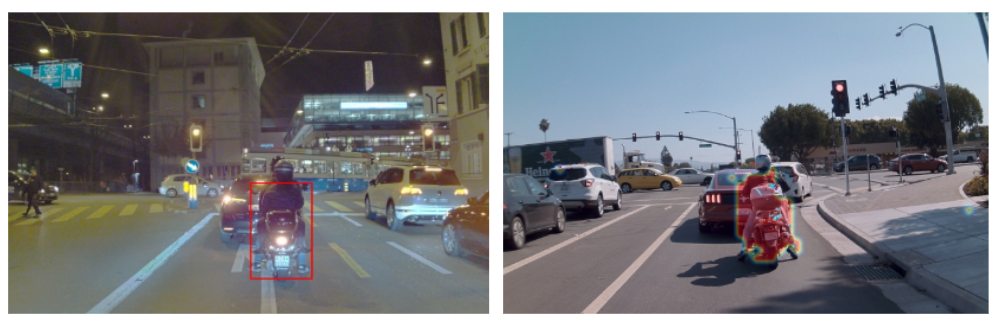}
    \caption{Heat map visualization of $M_Q'$ score map. \textbf{Left:} Input query image with region of interest \textbf{Right:} Top-1 semantically similar image using DTM with overlaid heat map.}
    \label{fig:mq_score_map}
\end{wrapfigure}


A larger $S(F_S, F'_Q)[x, y, i, j]$ value indicates that sample patch feature $F_S[x, y]$ and query patch feature $F'_Q[i, j]$ are more similar. Note that the tensor $S[x,y,:,:]$ stores similarity scores between sample patch feature $F_S[x, y]$ and all query patch features, and $S[:,:,i,j]$ stores similarity scores between query patch feature $F'_Q[i, j]$ and all sample patch features. 

Computing $S$ is highly efficient due to the following reasons: 1) Both $F'_Q$ and $F_S$ are L2 normalized which means that $|F_S[x, y]|_2 \cdot |F'_Q[i, j]|_2 = 1$. Hence, the cosine similarity in \equref{eq:simcomp}  boils down to computing a dot product along the channel dimension which is highly parallelizable on the GPU/CPU using off the shelf operations\footnote{See {\fontfamily{qcr}\selectfont tensorflow.tensordot, tensorflow.math.re- duce\_max, tensorflow.math.reduce\_mean}}  and 2) $F'_Q$ has all zeros except the ROI which makes it sparse thereby enabling the advantages of sparse matrix multiplications.

\subsection{Scoring Method}
For computing the score of the best matching sample patch feature for each query patch feature, we use a max over the cosine similarity scores stored in $S$. This gives a score map $M_Q' \in \mathbb{R}^{w \times h}$ which can be computed with



\begin{equation} \label{eq:m_q}
    M_Q'= \frac{\underset{axis=(0,1)}{\max}S(F_S, F'_Q)}{ 
A_Q'}
\end{equation}
where, $A_Q' \in \mathbb{R}^{w \times h}$ is a normalization constant that is proportional to the area of the individual projected ROI(s), i.e each spatial location in $A_Q'$ has a positive value equal to the area of its corresponding ROI in $F'_Q$ (if any). 
$M'_Q$ can be biased towards objects with larger ROIs in cases where the query image template has multiple ROIs with objects of different sizes occupying different ROI areas. We counter this with the normalization constant $A_Q'$.
Each element of $M_Q’[i, j]$, where $i \in  [0,w-1]$ and $j \in [0, h-1]$, indicates the best matching score found between the patch feature $F_Q’[i, j]$ and any patch feature in $F_S$, normalized by the area of the ROI at spatial location $(i, j)$, which is denoted by  $A_Q’[i, j]$ .


Using the score map $M_Q'$, we can compute the final score between query image $I_Q$ and sample image $I_S$ by averaging over the best patchwise similarity scores as
\begin{equation} \label{eq:sc1}
    Score(I_Q, I_S) = Mean(M_Q')
\end{equation}



Note that in $Score(I_Q, I_S)$ in \equref{eq:sc1}, we are not exploiting spatial relationships across features within the template. Computing scores that exploit spatial relationships explicitly tend to have higher runtime complexity \cite{talker2018efficient, talmi2017template, dekel2015best} while not yielding significant retrieval recall gains.
Our score map $M_Q'$ accurately matches the semantics of the ROI in the query image to the object in the sample image. Figure \ref{fig:mq_score_map} depicts patches that are being matched from the query image to the sample image. 

\section{Experiments} \label{sec:results}

\subsection{Experimental Setup} \label{sec:4.1}

We use internal research datasets for our experiments - one with 465k labeled images and another with 2.2M unlabeled images. We use a small representative set of 36 query images where the ROI in each query image is a motorcycle. These query motorcycles are chosen to have diverse characteristics like pose, size, orientation, lane location etc.. For fairness, we ensure disjoint driving sessions of the dataset and the query images. We use a one-stage object detector based on a UNet-backbone that was pre-trained on 900k labeled images to detect the classes: 'car', 'truck', 'pedestrian', 'bicycle' and 'motorcycle'. 
In our method, we represent an image by extracting features $F \in \mathbb{R}^{w \times h \times c}$ from the penultimate layer of this pre-trained object detector DNN. To reduce storage costs, we downsize the embedding by adding a maxpool layer. For comparison with the baseline approach \cite{babenko2015aggregating}, we represent an image by flattening $F$ using global average pooling such that $F \in \mathbb{R}^c$. In the following sections, we evaluate the performance of DTM on various settings.

\begin{wrapfigure}{R}{0.50\textwidth}
    \centering
    \includegraphics[width=0.50\textwidth]{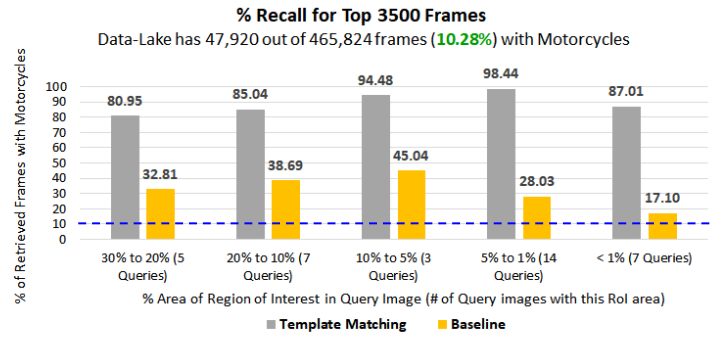}
    \caption{Top 3500 retrievals for each of the 36 query images, i.e., 126K retrievals. Relative gain of DTM Recall over Random (dotted line) is at least \textbf{8x} and over Baseline is at least \textbf{2x}.}
    \label{fig:3500_recall}
    \vspace{-0.5cm}
\end{wrapfigure}

\subsection{DTM Evaluation on Labeled Data} \label{sec:4.2}
In this set of experiments, we evaluate our DTM approach and compare it with the methodology proposed in \cite{babenko2015aggregating} (called baseline in \figref{fig:3500_recall}). The task is to find semantically similar motorcycles in a labeled dataset with 465k images.
This dataset has 10.28\% images with at least one motorcycle, which we use as an upper bound for random selection. 
We evaluate both the methods using Top-N recall where we count a mined image as a true positive if it has at least one motorcycle. 

\begin{figure*}
    \centering
    \includegraphics[width=1.0\textwidth]{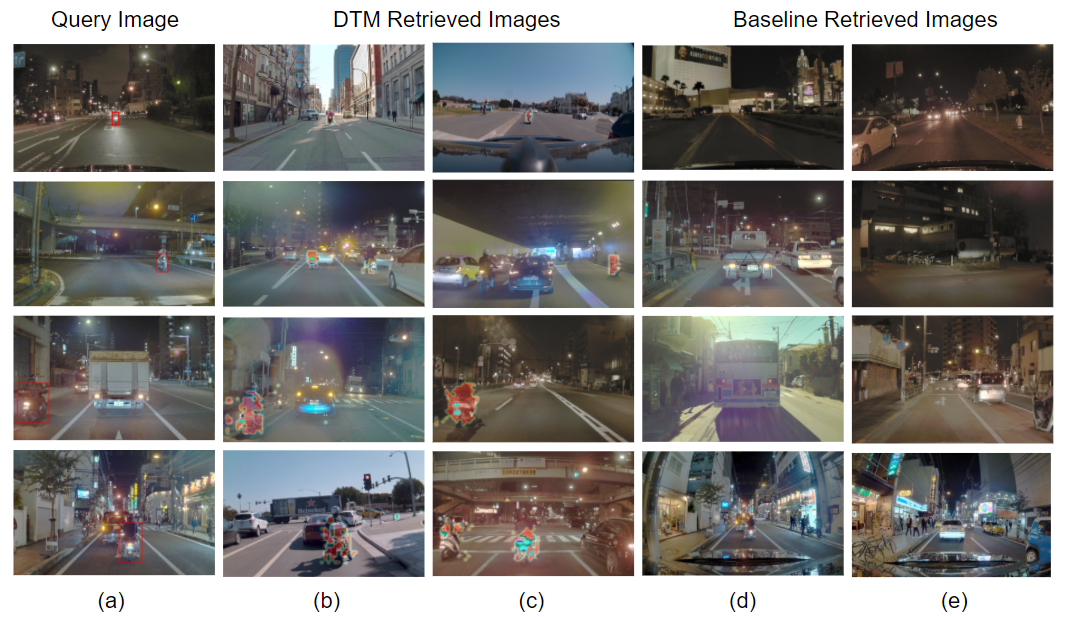}
    \caption{Results on our AV data: (a) Query image with region of interest (motorcycle). (b-c) DTM retrieved images containing motorcycles with similar semantics. (d-e) Baseline \cite{babenko2015aggregating} retrieved images that use semantics of the whole query image. Note how accurate the DTM heap maps are even for small objects with $\leq 1\%$ ROI area, e.g., the query image ROI in the top row.}
    \label{fig:labeled_exp}
    \vspace{-0.5cm}
\end{figure*}

Figure \ref{fig:3500_recall} shows the recall scores on  Top-3500 mined images. For fine-grained analysis, we divide the query set into 5 bins with respect to the area of the ROI in the query image. DTM (gray) outperforms baseline (yellow) and random by a significant margin. We can see that DTM has good recall across the entire range of the query ROI area. Whereas the recall for the baseline approach degrades significantly for $\leq 5\%$ ROI area (small objects, e.g., ROI in top row query image in \figref{fig:labeled_exp}). This is due to the fact that DTM does a patchwise similarity only using the ROI area from the query image. Figure \ref{fig:labeled_exp} demonstrates the qualitative results of DTM and the baseline. Our method produces sharp and localized heatmaps denoting its ability to accurately mine semantically similar objects. Since the baseline method uses a globally averaged flat embedding it cannot focus on the ROI. Hence, it retrieves images that represent the dominant objects/scenes in the query (e.g, urban scenes in the fourth row query, truck in the third row query).

\begin{wrapfigure}{R}{0.45\textwidth}
    \centering
    \includegraphics[width=0.45\textwidth]{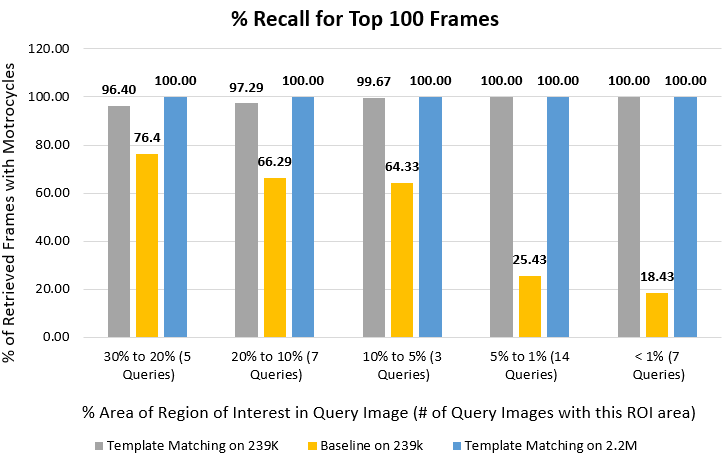}
    \caption{Visual inspection of top 100 retrievals for each of the 36 query images. Relative gain of DTM Recall over Random (3.2\%) is at least \textbf{32x} and over Baseline is at least \textbf{1.2x}.}
    \label{fig:100_recall}
    \vspace{-0.7cm}
\end{wrapfigure}

\subsection{DTM Evaluation on Unlabeled Data} \label{sec:4.3}
We use the same experimental setup described in section \ref{sec:4.1}  and \ref{sec:4.2} for evaluating DTM on a large unlabeled dataset with 2.2M images.  
We further shortlist a pool of 239k night-time images from the 2.2M images using metadata that indicates high likelihood for presence of at least one pedestrian or bicycle or motorcycle object in the image. 

We first evaluate DTM and baseline on the 239k unlabeled dataset, and next DTM on the 2.2M unlabeled dataset. For evaluation, we compute recall by visually inspecting the top 100 images mined for each of the 36 queries using the DTM and baseline method. A mined image is counted as a true positive if there exists at least one motorcycle in it. We observe consistent results and similar conclusions
for unlabeled data as in section \ref{sec:4.1} (see Fig. \ref{fig:100_recall}).

\subsection{Hard In-Distribution Queries} \label{sec:4.4}

\begin{wrapfigure}{R}{0.40\textwidth}
    \centering
    \includegraphics[width=0.40\textwidth]{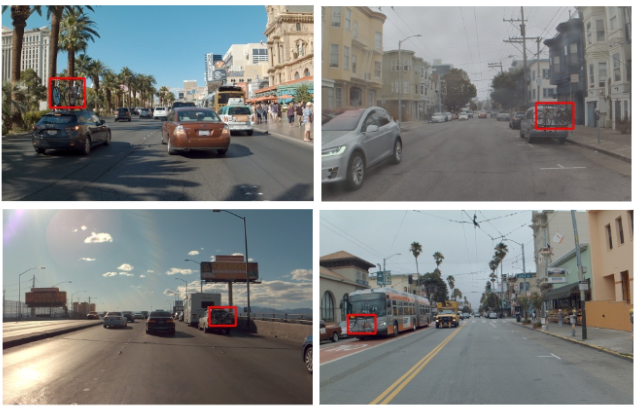}
    \caption{Bicycle hard queries: Various poses of bicycles mounted on vehicles. They are false negative failure cases - unusual and under-represented instances - that are interesting for mining.}
    \label{fig:hard_queries}
\end{wrapfigure}

In this experiment, we evaluate the performance of DTM in mining for hard queries. These hard queries typically have objects with unusual pose, size, occlusion or position in the image and tend to be under-represented or absent in the training dataset. Figure \ref{fig:hard_queries} shows an example of hard queries where a bicycle is mounted on the top/back/front of a vehicle. These variations of mounted bicycles were under-represented in the training dataset and were not detected by our object detector (false negatives).

\begin{figure*}
    \centering
    \includegraphics[width=1.0\textwidth]{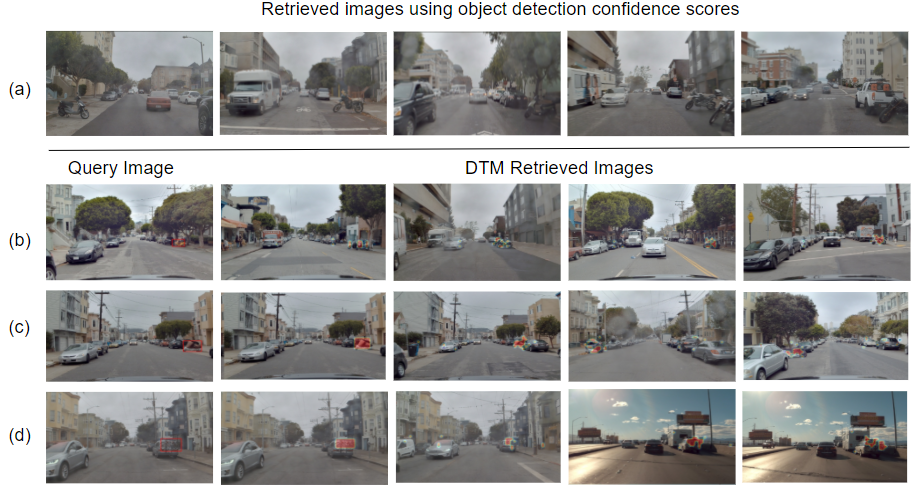}
    \caption{Comparison of DI (a) with DTM (b-d) for retrieving semantically similar motorcycles/bicycles. The DI method retrieves images independent of the query image. Notice DTM retrieving very similar images for hard failure cases like (b) side pose motorcycle (c) motorcycle covered in cloth (d) bicycle mounted on the back of a car.}
    \label{fig:exp2}
    \vspace{-0.7cm}
\end{figure*}

For these hard queries, we compare the performance of DTM with an image retrieval strategy using the inference of our object detector model. For ranking the images using the object detector model inference, we score each image as max of confidence values across all detected motorcycle objects in that image. The score is 0 if no motorcycle is detected. We refer to this as the detection inference (DI) method. We use max so that the top retrievals using DI contain a motorcycle with high probability. Note that the features used in DTM are derived from the same trained model used in DI. 

Figure \ref{fig:exp2} shows the visual comparison between DTM and DI. The DI method fails to mine for the semantics of the query ROI beyond the object category (see row (a) in \figref{fig:exp2}). This is undesirable for our task where images semantically similar to the failure case need to be mined. To validate this, we use DTM to retrieve images similar to a motorcycle/bicycle that the object detection model was not able to detect (false negative). We consider 19 motorcycle/bicycle hard query images and retrieve 20 images for each query. \textit{51\%} of these retrieved images turned out to be false negatives when inference was run on them, indicating that the retrieved images were also hard examples for the object detector model. Rows (b-d) in \figref{fig:exp2} show the top retrievals where we indeed see that DTM retrieves images with objects semantically similar to the failure case. In (c), DTM retrieves motorcycles covered in cloth and in (d), bicycles mounted on the back of a car.

\vspace{-3ex}
\subsection{Out-of-Distribution Queries} \label{sec:4.5}
\begin{figure*}
    \centering
    \includegraphics[width=1.0\textwidth]{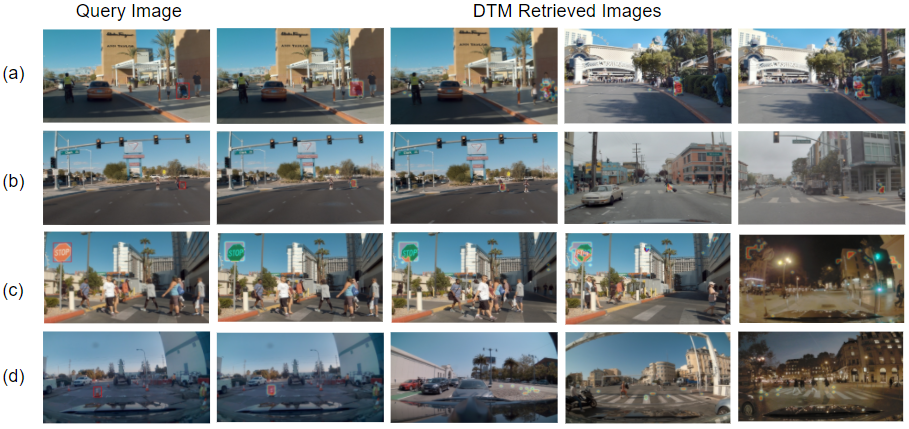}
    \caption{DTM retrieved images for out-of-distribution queries. (a) Stroller (b) Luggage (c) Stop sign (d) Traffic cone. In (a-b), DTM is able to retrieve semantically similar images from the same scene as the query as well as other scenes. For (c-d), DTM only retrieves images from the same scene since features for those objects are not modeled in the underlying DNN feature representation.}
    \label{fig:exp_ood}
\end{figure*}

We test our DTM algorithm for retrieving images that are semantically similar to objects that the underlying DNN was \textit{not} trained for. 
We observe that DTM performs well when the query is spatially co-located with objects that the object detector was trained for. For instance, in \figref{fig:exp_ood} (a-b), the query image has a stroller and luggage, both being dragged by a pedestrian. The object detector was trained on pedestrian but not on stroller or luggage. 
For totally unrelated objects, DTM fails to retrieve images with the object(s) of interest (see \figref{fig:exp_ood} (c-d) where the query is a stop sign and a traffic cone). However this is expected since the object detector from which the features were extracted was not trained on these objects (stop sign and traffic cone).
Many possible directions can be tried, like extracting the low-level features from an earlier layer of the DNN but we leave this for future research. 

\subsection{DTM for Multiple Regions of Interest} \label{sec:4.6}
\begin{figure*}
    \centering
    \includegraphics[width=1.0\textwidth]{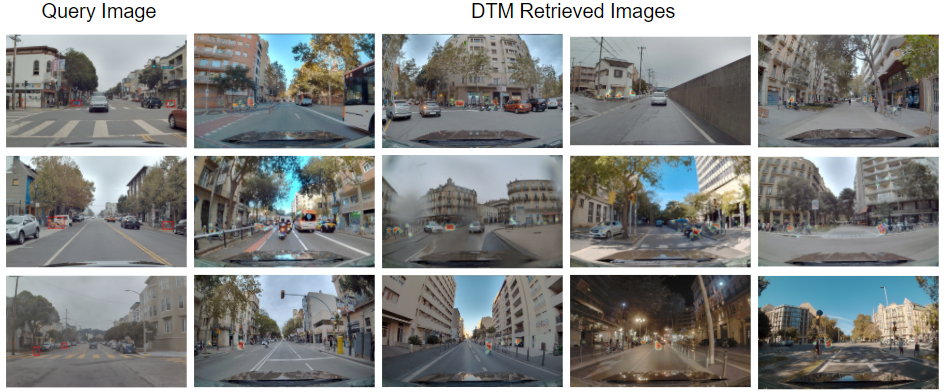}
    \caption{DTM retrieved images with multiple objects belonging to different semantic categories (motorcycle \textit{and} bicycle) co-occurring in the same sample image.}
    \label{fig:exp3}
\end{figure*}

DTM can be easily extended to accommodate queries with multiple regions of interest at no extra cost. This can be done by projecting all the ROIs in the query image onto the query feature map (as described in \figref{fig:dtm_arch}). We normalize the score maps using the area of each object so that the retrieved images are not biased towards larger objects (see \equref{eq:m_q}). For our AV use case, we evaluate the performance of DTM for retrieving semantically similar images with motorcycle \textit{and} bicycle. We achieve an average Top-100 recall score of \textit{88.5\%} across 4 queries, each with a pair of motorcycle and bicycle as ROIs. Figure \ref{fig:exp3} illustrates the qualitative performance of DTM for this task.

\section{Conclusion}
We propose a novel approach for object level targeted selection by using deep template matching. We study this problem in the context of mining images that are semantically similar to failure cases, like false negatives/positives of object detectors deployed in autonomous vehicles. These failure cases typically have unusual characteristics in terms of scale, pose, occlusion and tend to be absent or under-represented in the training dataset. Fixing these failure cases often involve mining for semantically similar objects from a large pool of unlabeled data. Our method focuses on the semantics of the objects of interest by projecting it onto the feature space and has high recall even when the object is small-sized, amid occlusion and heavy clutter. Our method works for multiple co-occurring objects in one or more semantic categories for object-level retrieval. Unlike other methods, it does not require extra labeled training data. 

\bibliography{main}

\end{document}